\documentclass[11pt,a4paper]{article}
\usepackage[hyperref]{acl2020}
\usepackage{times}
\usepackage{latexsym}
\usepackage{tikz}
\usepackage{microtype}
\usepackage{amssymb}
\usepackage{amsmath}
\usepackage{booktabs}
\usepackage{enumitem}
\usepackage{tabularx}

\usetikzlibrary{backgrounds}
\usetikzlibrary{decorations.pathreplacing}
\usetikzlibrary{patterns}
\usepackage{xspace}
\newcommand{\bfwn}{Bf+WN\xspace}
\newcommand{\atwn}{At+WN\xspace}
\newcommand{\atun}{At+Un\xspace}
\newcommand{\atin}{At+In\xspace}

\aclfinalcopy
\def\aclpaperid{1}

\title{Challenges in Emotion Style Transfer:\\ An Exploration with a
  Lexical Substitution Pipeline}

\author{David Helbig, Enrica Troiano \and Roman Klinger \\
  Institut f{\"u}r Maschinelle Sprachverarbeitung\\ University of Stuttgart, Germany \\
  \texttt{\{david.helbig,enrica.troiano,roman.klinger\}@ims.uni-stuttgart.de}\\
}

\date{}

\begin{document}
\maketitle
\begin{abstract}
  We propose the task of emotion style transfer, which is particularly
  challenging, as emotions (here: anger, disgust, fear, joy, sadness,
  surprise) are on the fence between content and style. To understand
  the particular difficulties of this task, we design a transparent
  emotion style transfer pipeline based on three steps: (1)~select the
  words that are promising to be substituted to change the emotion
  (with a brute-force approach and selection based on the attention
  mechanism of an emotion classifier), (2)~find sets of words as
  candidates for substituting the words (based on lexical and
  distributional semantics), and (3) select the most promising
  combination of substitutions with an objective function which
  consists of components for content (based on BERT sentence
  embeddings), emotion (based on an emotion classifier), and fluency
  (based on a neural language model). This comparably straight-forward
  setup enables us to explore the task and understand in what cases
  lexical substitution can vary the emotional load of texts, how
  changes in content and style interact and if they are at odds. We
  further evaluate our pipeline quantitatively in an automated and an
  annotation study based on Tweets and find, indeed, that simultaneous
  adjustments of content and emotion are conflicting objectives: as we
  show in a qualitative analysis motivated by Scherer's emotion
  component model, this is particularly the case for implicit emotion
  expressions based on cognitive appraisal or descriptions of bodily
  reactions.
\end{abstract}

\section{Introduction}
\label{sec:intro}
Humans are capable of saying the same thing in many ways. Careful
lexical choices can re-shape a concept in different modes of
presentation, giving it a humourous tone, for example, or some degree
of formality, or a rap vibe. This type of linguistic creativity has
recently been mirrored in the task of textual style transfer, where a
stylistic variation is induced on an existing piece of text. The core
idea is that texts have a content and a style, and that it is possible
to keep the one while changing the other.

\begin{figure}[b]
\setlength{\tabcolsep}{2pt}
\centering\footnotesize
\begin{tabular}{|lccccc|}
\hline
In (Anger):&This & \textbf{soul-crushing} & \textbf{drudgery}& \textbf{plagues}& him\\
Out (Joy): &This & \textbf{fulfilling} & \textbf{job} & \textbf{motivates} &him\\
\hline
\end{tabular}
\caption{An example of emotion transfer performed with lexical substitution.}
\label{fig:task_example}
\end{figure}

Past work on style transfer has targeted attributes (or styles) like
sentiment \cite{Dai2019} and tense \cite{Hu2017}, producing a rich
literature on deep generative models that disentangle the content and
the style of an input text, and subsequently condition generation
towards a desired style \cite{Fu2018,Shen2017,Prabhumoye2018}. With
this paper, we propose a non-binary style transfer setting, namely
emotion style transfer, in which the target corresponds to one emotion
(following Ekman's fundamental emotions of anger, fear, joy, surprise,
sadness, and disgust). Further, this setting is particularly
challenging as emotions are on the fence between content and style. To
the best of our knowledge, this type of attribute has been explored
only to some degree by the unpublished work by \newcite{Smith2019},
who transfer text towards 20 affect-related styles. Emotions received
more attention in conditioned text generation
\cite{Ghosh2017,Huang2018,Song2019}.

To explore the challenges of emotion style transfer (for which we
depict an example in Figure \ref{fig:task_example}), we develop a
transparent pipeline based on lexical substitution (in contrast to a
black-box neural encoder/decoder approach), in which we first (1)
select those words that are promising to be changed to adapt the
target style, (2) find candidates that may substitute these words, (3)
select the best combination regarding content similarity to original
input, target style, and fluency. As we will see, this
straight-forward approach is promising while it still enables to
understand the changes to the text and their function.

Emotions are not only interesting from the point of view that they
contribute to content and style. They are also a comparably
well-investigated phenomenon with a rich literature in psychology. For
instance, \newcite{Scherer2005} states that emotions consist of
different components, namely a cognitive appraisal, bodily symptoms, a
subjective feeling, expression, and action tendencies. Descriptions of
all these components can be realized in natural language to
communicate a specific private emotional state. We argue (and analyze
based on examples later) that a report of a feeling (``\emph{I am
  happy}'') might be challenging in a different way than descriptions
of bodily reactions (\emph{``I am sweating''}) or events (\emph{``My
  dog was overrun by a car''}).

With our white-box approach of style transfer and the evaluation on
the novel task of emotion transfer, we address the following research
questions: To what extent can lexical substitution modulate the
emotional leaning of text? What is its limitation (e.g., by changing
the emotion ``style'', does content change as well)? Our results show
that the success of this approach, both in terms of style change and
content preservation, depends on the strategies used for selection and
substitution, and that emotion transfer is a viable task to
address. Further, we see in a qualitative analysis that what an
emotion classification model bases its decisions on might not be
sufficient to guide a style transfer method. This becomes evident when we 
compare how transfer is realized across types of emotion expressions, 
corresponding to specific components of Scherer's model.

Our implementation is available at
\url{http://www.ims.uni-stuttgart.de/data/lexicalemotiontransfer}.

\section{Related Work}

\subsection{Emotion Analysis}
In the field of psychology, the two main emotion traditions are
categorical models and the strand that focuses on the continuous nature
of humans' affect \cite{Scherer2005}. Emotions are grouped into
categories corresponding to emotion terms, some of which are
prototypical experiences shared across cultures. For
\newcite{ekman_argument_1992}, they are anger, joy, surprise, disgust,
fear and sadness; on top of these, \newcite{plutchik_nature_2001} adds
anticipation and trust. \newcite{Posner2005}, instead locates emotions
along interval scales of affect components (valence, arousal,
dominance).

These studies have also influenced computational approaches to
emotions, whose preliminary requirement is to follow a specific
conceptualization coming from psychology, in order to determine the
number and type of emotion classes to research in language.  Emotion
analysis in natural language processing has mainly established itself
as a classification task, aimed at assigning a text to the emotion it
expresses \cite{alm_emotions_2005}. It has been conducted on a variety
of corpora that encompass different types of annotations\footnote{A
  comprehensive list of available emotion datasets and annotation
  schemes can be found in \newcite{Bostan2018}.}, based on
one of the established emotion models mentioned above. Such studies
also differ with respect to the textual genres they consider, ranging
from from tweets \cite{Mohammad2017c,Klinger2018} to literary texts
\cite{Kim2017a}.

While emotion classification approaches have been used to
guide controlled generation of text
\cite{Ghosh2017,Huang2018,Song2019}, computationally 
modelling emotions has not yet been applied to style
transfer. After describing a method to address such task, we analyse 
its performance by leveraging Scherer's component model: emotions are 
underlied by various dimensions of cognitive appraisal, which can be differently 
expressed in text and may pose different challenges for style transfer.

\begin{figure*}
  \centering
  \fbox{\includegraphics[scale=1.1]{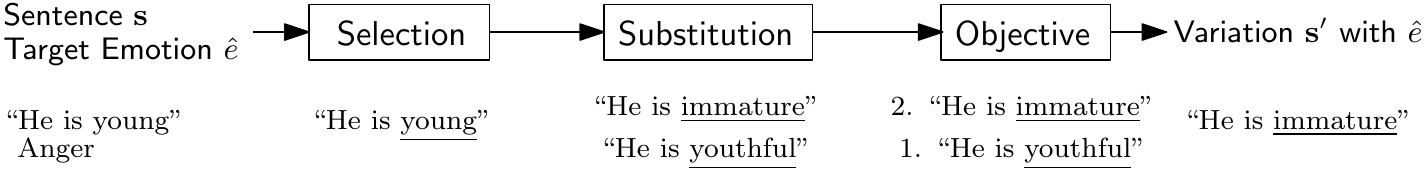}}
  \caption{Pipeline model architecture. The selection module marks
    tokens to substitute, the substitution module retrieves candidates and perform
    substitution. The objective ranks and scores variations.}
\label{fig:pipeline}
\end{figure*}

\subsection{Style Transfer}
Most of the recently published approaches to style transfer make use of
artificial neural network architectures, in which some latent semantic
representation is the backbone of the system. For instance,
\newcite{Prabhumoye2018} use neural back-translation to encode
the content of text while reducing its stylistic properties, and later
decoding it with a specific target
style. \newcite{gong_reinforcement_2019} evaluate paraphrases
regarding their fluency, similarity to the input text and expression
of a desired target style, and use this as feedback in a reinforcement
learning approach. \newcite{li_delete_2018} combine rules with neural
methods to explicitly encode attribute markers of the target style.

Such transfer methods have been applied to a variety of styles,
including sentiment \cite{Shen2017,Fu2018,Xu2018} and a number of
affect-related variables \cite{Smith2019}. Other examples include text genres
\cite{lee_neural_2019,jhamtani-etal-2017-shakespearizing}, romanticism
\cite{li_delete_2018}, politeness/offensiveness and formality
\cite{sennrichetal2016,nogueira-dos-santos-etal-2018,wang-etal-2019-harnessing}.

One of the earliest methods that targets sentiment is proposed by
\newcite{guerini_valentino:_2008}, who change, add and delete
sentiment-related words in a lexical substitution framework. Their
strategy to retrieve candidate substitutes is informed by a thesaurus
and an emotion dictionary: the first facilitates the extraction of
substitutes standing in a specific semantic relation to the input
words, the other allows to pick those words that have the desired
valence score. Following this approach,
\newcite{whitehead_generating_2010} filter out ungrammatical
expressions resulting from lexical substitution.

Like some works mentioned above, we adopt the view that emotions can
be transfered by focusing on specific words, we use WordNet as a
source of lexical substitutes, and we consider the three objectives of
fluency, similarity and the presence of the target style. Moreover, we
opt for a more interpretable solution than neural strategies, as we
aim at pointing out what leads to a successful transfer, and what, on
the contrary, prevents it.

\subsection{Paraphrase Generation through Lexical Substitution}
Lexical substitution received some attention independent of style
transfer, as it is useful for a range of applications, like paraphrase
generation and text summarisation \cite{dagan-etal-2006-direct}.  This
task, which was formulated by \newcite{mccarthy-navigli-2007-semeval}
and implemented as part of the SemEval-2007 workshop, consists in
finding lexical substitutes close in meaning to the original word,
given its context within a sentence. The task has mainly been
addressed using handcrafted and crowdsourced thesauri, such as
WordNet, in order to retrieve lexical substitutes
\cite{martinez-etal-2007-melb,sinha2014explorations,kremer-etal-2014-substitutes,biemann2013creating}. Moreover,
it has been approached with distributional spaces, where the
embeddings of the candidate substitutes of a target word can be found,
and they can be ranked according to their similarity to the target
embedding \cite{zhao-etal-2007-hit,hassan-etal-2007-unt}, as well as the
similarity of their contextual information
\cite{melamud_simple_2015}\footnote{A comparison of different
  context-aware models for lexical substitution can be found in
  \newcite{soler_comparison_2019}.}.

In the present paper, we follow a similar progression: we retrieve
candidates for lexical substitution in WordNet; then, in our more
advanced systems, we switch to embedding-based retrieval models.

\section{Methods}

Emotion transfer can be seen as a task in which a sentence
$\mathbf{s}$ is paraphrased, and the result of this operation exhibits
a different emotion than $\mathbf{s}$, specifically, a target
emotion. We address emotion transfer with a pipeline in which each
unit contributes to the creation of emotionally loaded
paraphrases. The pipeline is shown in Figure~\ref{fig:pipeline}. First
is a \textit{selection} component, which identifies the tokens in
$\mathbf{s}$ that are to be changed. Then, the \textit{substitution}
component takes care of the actual substitution. It is responsible for
finding candidate substitutes for the tokens that have been selected,
producing paraphrases of the input sentence. Importantly, paraphrases
are over-generated: at this stage of the pipeline, the output is
likely to include sentences that do not express the target
emotion. Paraphrases are then scored and re-ranked in the last,
\text{objective} component, which picks up the ``best'' output.

\subsection{Selection}
This component identifies those tokens from a sentence
$\mathbf{s}=t_1,\ldots t_n$ that will be substituted later, and groups
them into \textit{selections} $\mathcal{S}=\{S_i\}$, where each $S_i$
consists of tokens, $S_i=\{t_i,\ldots,t_j\}$ ($1\geq i,j\leq n$). We
experiment with two selection strategies, in which the maximal number
of tokens in one selection is $p$ and the maximal number of selections
is $q$ ($p,q \in\mathbb{N}$).

\paragraph{Brute-Force.} This baseline selection strategy picks each
token separately, therefore, we obtain $n$ selections, one for each
token, i.e., $\mathcal{S} = \{\{t_1\}, \ldots, \{t_n\}\}$ ($p=1$,
$q=n$).

\paragraph{Attention-based.}
To pick words that are likely to influence the (current and target)
emotion of a sentence, we exploit an emotion classification model to
inform the selection strategy. We train a bi\-LSTM with a
self-attention mechanism \cite{baziotis_ntua-slp_2018} and then select
those words with a high attention weight to be in the set of
selections. To avoid a combinatorial explosion, we consider the $k$
tokens with highest attention weights and add all possible
combinations of up to $p$ tokens. Therefore, $q=|\mathcal{S}|=
\sum\nolimits_{i=1}^k \binom{p}{i}$. As an example, possible
selections in the sentence from Figure~\ref{fig:task_example} for $k =
3, p=2$ would be $S = \{ \{\textrm{soul-crushing}\},\allowbreak
\{\textrm{drudgery}\},\allowbreak \{\textrm{plagues}\},\allowbreak
\{\textrm{soul-crushing, drudgery}\},\allowbreak
\{\textrm{soul-crushing, plagues}\},\allowbreak \{\textrm{drudgery,
  plagues}\}\}$.

\subsection{Substitution}
The selections $\mathcal{S}$ are then passed to the substitution model
together with part-of-speech information. Two tasks are fulfilled by
this component: substitution candidates are found for the tokens of
each $S_i$, and the substitution is done by replacing those candidate
tokens at position $i,\ldots, j$ in the input sentence
$\mathbf{s}$. The next paragraphs detail our strategies for candidate
retrieval. We compare a lexical semantics and two distributional
semantics-based methods.

\paragraph{WordNet Retrieval.}
In the WordNet-based method \cite{fellbaum_wordnet:_1998}, we retrieve
the synsets for the respective selected token with the assigned part
of speech. Candidates for substitution are the neighboring synsets
with the hyponym and hypernym relation (for verbs and nouns) and
antonym and synonym relation (for adjectives).

Note that we do not perform word-sense disambiguation prior to
retrieving the base synsets. Accordingly, the sense of the selected
token in the context of the source sentence and the sense of some
retrieved candidates may be different. This is in line with the design
of the pipeline and we expect irrelevant forms to be penalised in the
objective component.

\paragraph{Distributional Retrieval -- Uninformed.} 
In the ``Distributional Retrieval -- Uninformed'' setting, we retrieve
$u$ substitution candidates based on the cosine similarity in a vector
space.  To build the vector space, we employ pre-trained word
embeddings.\footnote{300 dimensional embeddings, available at
  \url{https://github.com/cbaziotis/ntua-slp-semeval2018}} They are
the same that are used for training the emotion classifier responsible
for retrieving attention scores in the selection stage.

\paragraph{Distributional Retrieval -- Informed.} 
A disadvantage of the uniformed method mentioned before might be that
the selected $u$ substitutions for each token might not contain words
with the targeted emotional orientation. In this approach, we slightly
change the substitution selection process by first retrieving a list
of $u$ most similar tokens from the vector space. Based on this list,
which is presumably of sufficient similarity to the selected token, we
select those $v$ relevant for the target emotion.

Let $E$ be the set of emotion categories and $\hat{e}\in E$ the target
emotion (with vector representation $\hat{\mathbf{e}}$). Further, let
$\bar{\mathbf{e}}$ be the centroid of concepts associated with the
respective emotion, as retrieved from the NRC emotion dictionary
\cite{mohammad_crowdsourcing_2013}. From the list of semantically
similar $u$ candidates $c$ for one token to be substituted, we select
the $v$ top scoring ones via
\[
\textrm{score}(c,\hat{e}) = 
  \cos(\hat{\textbf{e}},\mathbf{c}) - 
    \frac{1}{|E|-1}
     \sum_{\bar{e}\in E\backslash \hat{e}}
       \cos(\bar{\mathbf{e}},\mathbf{c})\,.
\]

\subsection{Objective}
\label{sec:objective}
The set of candidate paraphrases produced at substitution time, based on the
selections, are an overgeneration which might not be fluent, diverge
from the original meaning, and might not contain the target
emotion. To select those paraphrases which do not have such unwanted
properties, we subselect those with the desired properties based on an
objective function $f(\cdot)$ which consists of three components for
fluency of the paraphrase $\mathbf{s}'$,
semantic similarity between the original sentence $\mathbf{s}$ and the
paraphrase $\mathbf{s}'$, and the target emotion $\hat{e}$ of the
paraphrase, therefore
\[
\textrm{f}(\mathbf{s}\kern-1pt,\kern-1pt\mathbf{s}'\kern-4pt,\hat{e})\kern-1pt =\kern-1pt\\
\lambda_{1} \cdot \textrm{emo}(\mathbf{s}',\hat{e}) +
\lambda_{2} \cdot \textrm{sim}(\mathbf{s},\mathbf{s}') +
\lambda_{3} \cdot \textrm{flu}(\mathbf{s}').
\]
The paraphrase with the highest final score is selected as the result
of the emotion transfer process ($\sum_i \lambda_i=1$).

\paragraph{Emotion Score.}
To obtain a score for the target emotion $\hat{e}$ we use an emotion
classification model (the same as for the attention selection
procedure) in which the last layer is a fully connected layer of size $|E|$ and the output layer is a softmax.
Let $g$ represent the classification model that takes a sequence of tokens $\mathbf{s}$ and an emotion $e$ as inputs and produces the activation for $e$ in the final layer. Therefore,

\[
\textrm{emo}(\mathbf{s}',\hat{e}) = \frac{\exp(g(\mathbf{s}', \hat{e}))}{\sum_{e \in E} \exp(g(\mathbf{s}', e))}.
\]

\paragraph{Similarity Score.}
To keep the semantic similarity as much as possible between the input
sentence $\mathbf{s}$ and the candidate paraphrase $\mathbf{s}'$, we
calculate the cosine similarity between the respective sentence
embeddings, based on the pre-trained BERT model \cite{devlin-etal-2019-bert}, in the implementation
provided by \newcite{wolf_huggingfaces_2019}.
We conceptualize BERT as a mapping function that takes a sequence of tokens $\mathbf{s}$ as input and produces a hidden vector representation for each token.
The sentence embeddings $\mathbf{r}$ are obtained by averaging over all hidden vectors.%
\footnote{As recommended in the documentation of the implementation
by Wolf et al. (2019) (\url{https://huggingface.co/transformers/model_doc/bert.html}, accessed on March 27, 2020), we do not use the reserved classification token [CLS] as a sentence embedding.}
Therefore,
\[
 \textrm{sim}(\mathbf{s},\mathbf{s}')  = \cos(\mathbf{r}, \mathbf{r}').
\]

\paragraph{Fluency Score.} 
To avoid that tokens are substituted with words which do not fit in
the context, we include a language model which scores the paraphrase
$\mathbf{s}'$ \cite[similar to][]{zhao2018}. This model assesses the fluency
by perplexity using GPT
\cite{radford_improving_2018}, an autoregressive neural language model
based on the transformer architecture, which allows us to read the
probability of the next token in a sentence given its history.  We use
a pretrained version of the model provided by
\newcite{wolf_huggingfaces_2019}. The perplexity as
the average negative log probability over the tokens of our variation
sentence $\mathbf{s}'$ is
\begin{multline*}
\textrm{perplexity}(\mathbf{s}') =\\ \frac{1}{n-1} \sum_i^{n-1} -\log(P(t_{i+1}|t_1,\ldots,t_i)).
\end{multline*}
Since we are dealing with negative log values, a low perplexity score
indicates high probability and therefore high fluency.  In order to
obtain our final fluency score, we normalize the perplexity to the
range $[0,1]$ and reverse the polarity.  To this end, we use the
highest perplexity score ($\textrm{perplexity}_{\max}$) and lowest
perplexity score ($\textrm{perplexity}_{\min}$) that we retrieve among all
variation sentences created for our input sentence as scaling factors:
\[
\textrm{flu}(\mathbf{s}') = \frac{\textrm{perplexity}(\mathbf{s}')-\textrm{perplexity}_{\max}}{\textrm{perplexity}_{\min} - \textrm{perplexity}_{\max}}
\]

\begin{figure*}[t]
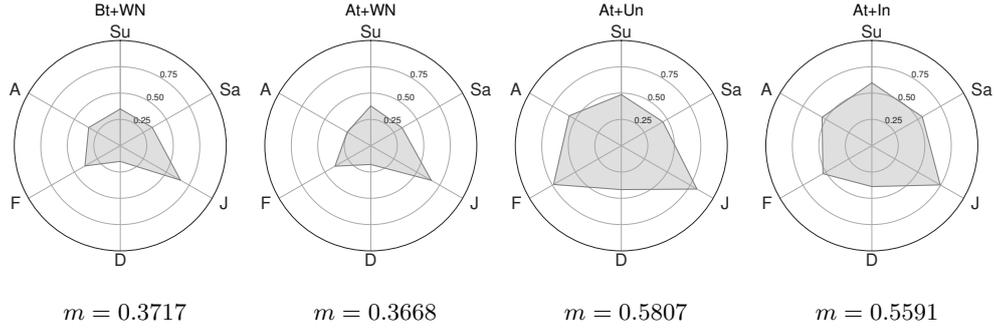

\centering
\begin{minipage}{.2\textwidth}
\resizebox{\textwidth}{!}{
\input{Bt+WN.pgf}}
\center{\small{$m = 0.3717$}}
\end{minipage}
\begin{minipage}{.2\textwidth}
\resizebox{\textwidth}{!}{
\input{At+WN.pgf}}
\center{\small{$m = 0.3668$}}
\end{minipage}
\begin{minipage}{.2\textwidth}
\resizebox{\textwidth}{!}{
\input{At+Un.pgf}}
\center{\small{$m = 0.5807$}}
\end{minipage}
\begin{minipage}{.2\textwidth}
\resizebox{\textwidth}{!}{
\input{At+In.pgf}}
\center{\small{$m = 0.5591$}}
\end{minipage}
\caption{Automated evaluation results. Each radar plot shows the
  average emotion scores achieved by transferring 1,000 tweets to
 anger (A), disgust (D), fear (F), joy (J), sadness (Sa) and 
 surprise (Su); $m$ is the average over all emotions.}
\label{fig:autoeval}
\end{figure*}

\section{Experiments}
Having established the general pipeline, we move on to the question
whether our strategies for selection and substitution actually produce
variations with the desired emotion (RQ1). In addition, we examine the
interaction between the emotion connotation of the paraphrases and
their similarity to the inputs (RQ2). These questions are answered in
an automatic and a human evaluation.

\subsection{Setting}
We instantiate and compare four model configurations for lexical
substitution with different combinations of selection and substitution
components. These are designed such that we can compare the selection
procedure separately from the substitution component.

\begin{itemize}[nosep,leftmargin=0em,labelwidth=*,align=left]
\item \textbf{\bfwn}: We select isolated words in the brute-force
  configuration and substitute those with the WordNet-based approach.
\item \textbf{\atwn}: To compare if the attention mechanism is more
  powerful in finding relevant words to be substituted, we change the
  brute force selection to the attention-based method. Here, we
  consider the tokens with the $k=2$ highest attention scores and
  combine them to selections with a maximum of $p=2$ tokens in each
  selection.
\item \textbf{\atun}: We keep the attention mechanism for selection
  with $k=2$ and $p=2$,
  but vary the substitution component to select $u=150$ candidates
  based on semantic similarity. As embedding space, we employ the same
  pre-trained embeddings we use for training the emotion classifier
  responsible for retrieving attention weights and calculating emotion
  scores.  The number of variations created amounts to $\sum_{i=1}^{p}
  \binom{k}{i} u^i = 2 \cdot 150 + 1 \cdot 150^2 = 22800$.
\item \textbf{At+In}: While the model configuration \atun generates
  many possibly irrelevant variations, this model makes informed
  decisions on how to substitute: we keep the selection as in \atun,
  but exchange the substitution method with the informed
  strategy. Specifically, $u=100$ candidates are found based on their
  semantic similarity to the token to be substituted, and among those,
  $v=25$ tokens are subselected based on their emotion-informed score,
  leading to $\sum_{i=1}^{p} \binom{k}{i} v^i = 3 \cdot 25 + 3 \cdot
  25^2 = 1950$ variations (with $k=3$, $p=2$). To inform this method
  about emotion in the embedding space, we use the NRC emotion
  dictionary \cite{mohammad_crowdsourcing_2013}.
\end{itemize}

\begin{figure}
  \centering
  \includegraphics{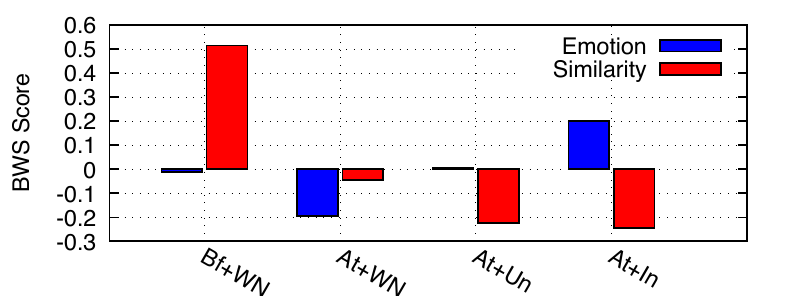}
  \caption{Results for the two human annotation trials, combined by model configuration.}
  \label{fig:humeval}
\end{figure}

\paragraph{Automatic Evaluation.}
The main goal of the automatic evaluation is to compare the potential
of increasing the probability that the paraphrase contains the target
emotion. To achieve that, we compare the four pipeline configurations,
but only use the emotion score as the objective function to pick the
best candidate. We use 1000 uniformly sampled Tweets from the corpus
TEC \cite{mohammad2012emotional}. The emotion classification model
used for scoring is trained on the same corpus using pre-trained
Twitter embeddings provided by \newcite{baziotis_ntua-slp_2018}.%
\footnote{\url{https://github.com/cbaziotis/ntua-slp-semeval2018}}.
We use the attention scores obtained from this model for our
attention-based selection method.
As embedding space for the \atun and \atin models, we use the same
embeddings. As we transfer to the six emotions annotated in TEC, we
obtain 6,000 paraphrases with \atun and \atin and 5,904 with \bfwn and
\atwn (the latter due to non-English words which are not found in
WordNet).

\paragraph{Human Evaluation.} The goal of the human evaluation is to
verify the automatic results (the potential of the selection and
substitution components). Further, we compare the association of the
paraphrase with the target emotion. To compare a basic setup and the
most promising setup, we use $\textrm{emo}(\textbf{s}',\hat{e})$ and
$\textrm{sim}(\textbf{s},\textbf{s}')$ for \bfwn, \atwn, and \atun and
$\textrm{flu}(\textbf{s}')$ in addition for At+In. This evaluation
is based on 100 randomly sampled Tweets for which we ensure that they
are single sentences from TEC. The annotation of emotion connotation
and similarity to the original text is then setup as a
best-worst-scaling experiment \cite{louviere_best-worst_2015}, in
which each of our two annotators is presented with one paraphrase for
each of the four configurations, all for the same emotion (randomly
chosen as well). Note that in contrast to best-worst scaling used for
annotation as, e.g., in emotion intensity corpus creation
\cite{mohammad_semeval-2018_2018}, where textual instances are scored,
here the instances change from quadruple to quadruple, but the
originating configurations remain the same and receive the score.  The
agreement calculated with Spearman correlation of both annotators is
$\rho=1$ for the emotion connotation and $\rho=0.8$ for semantic
similarity.

\begin{table}
  \centering\small
  \setlength{\tabcolsep}{10pt}
  \begin{tabularx}{\linewidth}{p{13mm}X}
    \toprule
    Text & Target \\
    \cmidrule(r){1-1}\cmidrule(r){2-2}
    Input & surprises are great when the person is surprised ! \\
    Output for Sadness & \textit{depresses} are great when the person is \textit{disappointed} ! \\
    \cmidrule(r){1-1}\cmidrule(r){2-2}
    Input & love watching my daughter be so excited around christmas \\
    Output for Anger & \textit{detest} watching my daughter be so \textit{annoyed} around christmas \\
    \bottomrule
  \end{tabularx}
  \caption{Examples of paraphrases produced with
    At+Inf for different target emotions, using all three components of the objective function. }
  \label{tab:best-configurationExamples}
\end{table}

\begin{table*}
\centering\small
\begin{minipage}{0.48\linewidth}
  \setlength{\tabcolsep}{2pt}
  \begin{tabular}{llll}
\toprule
ID & Text & Type & Target \\
\cmidrule(r){1-1}\cmidrule(r){2-2}\cmidrule(r){3-3}\cmidrule{4-4}
1  & \textbf{I am happy}                                 & Ex &   \\
   &  i \textit{fuck} \textit{annoyed}                   &   & A \\
   &  i \textit{dislike} \textit{crabby}                 &   & D \\
   &  i \textit{regret} \textit{king}                    &   & F \\
   &  \textit{and} am \textit{happy}                     &   & J \\
   &  i am \textit{bummed}                               &   & Sa \\
   &  i am \textit{surprise}                             &   & Su \\
\cmidrule(r){1-1}\cmidrule(r){2-2}\cmidrule(r){3-3}\cmidrule{4-4}
2  & \textbf{I am sad}                                   & Ex &   \\
   & i am \textit{angrier}                               &   & A \\
   & i \textit{embarrassed disgusting}                   &   & D  \\
   & i \textit{must lies}                                &   & F  \\
   & \textit{finally} am \textit{tiring}                 &   & J  \\
   & i \textit{depressed} sad                            &   & Sa \\
   & i \textit{came realise}                             &   & Su \\
\cmidrule(r){1-1}\cmidrule(r){2-2}\cmidrule(r){3-3}\cmidrule{4-4}
3  & \textbf{Tears are running over my face}              & BR &   \\
   & \textit{rage fuck} running over my face              &    & A \\
   & \textit{puke} are \textit{puking} over my face       &    & D \\
   & \textit{shadows} are \textit{creeping} over my face  &    & F \\
   & \textit{gladness} are running over my face           &    & J \\
   & \textit{depressed} are \textit{leaving} over my face &    & Sa\\
   & \textit{squealed came} running over my face          &    & Su\\
\bottomrule
\end{tabular}
\end{minipage}
\hfill
\begin{minipage}{0.48\linewidth}
  \setlength{\tabcolsep}{2pt}
  \begin{tabular}{llll}
\toprule
ID & Text & Type & Target \\
\cmidrule(r){1-1}\cmidrule(r){2-2}\cmidrule(r){3-3}\cmidrule{4-4}
4  & \textbf{I was tembling}                          & BR & \\
   & \textit{fuck irked} trembling                    &                 & A \\
   & \textit{fatass reeks} trembling                  &                 & D \\
   & i \textit{hallucinated trembling}                &                 & F \\
   & \textit{finally finally} trembling               &                 & J \\
   & \textit{bummed} was \textit{trembling}           &                 & Sa\\
   & \textit{mom showed} trembling                    &                 & Su\\
\cmidrule(r){1-1}\cmidrule(r){2-2}\cmidrule(r){3-3}\cmidrule{4-4}
5  & \textbf{My son was standing close to the street}               & Ap & \\
   & my \textit{fuck} was standing \textit{annoyed} to the street   &  & A\\
   & my \textit{molest} was \textit{peeing} close to the street     &  & D\\
   & my \textit{coward} was \textit{creeping} close to the street   &  & F\\
   & my \textit{yeshua} was \textit{soaking} close to the street    &  & J\\
   & my \textit{funeral} was \textit{leaving} close to the street   &  & Sa\\
   & my \textit{son} was standing \textit{surprise} to the street   &  & Su\\
\cmidrule(r){1-1}\cmidrule(r){2-2}\cmidrule(r){3-3}\cmidrule{4-4}
6  & \textbf{My grandmother died}                     & Ap &          \\
   & \textit{fckin} grandmother \textit{punched}      &                 & A  \\
   & \textit{ugh} grandmother \textit{farted}         &                 & D  \\
   & my \textit{voldemort attack}                     &                 & F  \\
   & my \textit{family rededicated}                   &                 & J  \\
   & \textit{cried} grandmother died                  &                 & Sa \\
   & my \textit{mama showed}                          &                 & Su \\
\bottomrule
\end{tabular}
\end{minipage}
\caption{Challenging cases for different ways to communicate an
  internal emotion state. Inputs are in bold; all paraphrases are produced with
  At+Inf and all three components of the objective function. Ex:
  Explicit emotion mention, BR: Bodily reaction, Ap: Event appraisal.}
\label{tab:outputExamples}
\end{table*}

\subsection{Results}
\paragraph{RQ1: Whats is the potential of emotion transfer with lexical substitution?}
\label{sec:RQ1}

We answer RQ1 by inspecting how likely the paraphrases are to contain the
desired emotion and first turn to the automatic evaluation. 
Figure~\ref{fig:autoeval} shows the results. Each radar plot indicates
the extent to which the paraphrases of each configuration express the
target emotions. The average probability of the
target emotion in the best paraphrases of \bfwn is 0.3717, indicating 
that this method has a slightly higher
potential than \atwn (0.3668); still, the shape of their plots is comparable.  When
we compare the substitution method while keeping the selection fixed
(\atwn, \atun, \atin), we see that the distributional methods show a
clear increase (0.5807 and 0.5591 average target emotion probability).

In the manual evaluation, we see in Figure~\ref{fig:humeval} (in
blue) that the results are in line with the automatic
evaluation. Instances originating from \atin are most often chosen 
as the best results, followed by \atun and \bfwn. \atwn scores the
worst in human evaluation. Note that the best-worst-scaling results
cannot directly be compared to automatic evaluation measures obtained
with an automatic text classifier.

\paragraph{RQ2: Is semantic content preserved when changing the emotional orientation?}
We answer this research question based on the human annotation
experiment, with the results in Figure~\ref{fig:humeval}. Contrary to
the results on the transfer potential, Bf is judged as the most
efficient selection strategy for content preservation, while At
configurations are dispreferred. The ones based on distributional
substitution appear to be worse compared to solutions leveraging
WordNet. This shows that Bf provides a lower degree of freedom to
the substitution component. The attention mechanism finds the relevant
words to be substituted, but the annotators perceive these
changes also as a change to the content.

\bigskip
\noindent
To sum up, highest transfer potential is reached with a combination of
attention-based selection, and distributional substitution. The fact
that the latter surpasses WordNet-based retrieval may be traced back
to the richness of embedding spaces, where substitution candidates can
be found which have a higher semantic variability than those found in
the thesaurus, and hence, have more varied emotional connotations. In
addition, the distributional strategy performing better is the
emotion-informed one (0.2 in Figure~\ref{fig:humeval}). This suggests
that accessing emotion information during substitution is beneficial.
The performance of this configuration is exemplified in
Table~\ref{tab:best-configurationExamples}, and further discussed in
the qualitative analysis.

By comparing the two human trials, it emerges that no configuration
excels in both emotion transfer and meaning preservation. In the
second case, Attention-based configurations are largely downplayed by
\bfwn. Therefore, to tackle RQ2, the more a system changes emotions,
the less it preserves content.

\section{Analysis}
We now turn to a more qualitative analysis of the results. Due to
space restrictions, we show examples for the four pipeline
configurations, all with the same objective function
$\textrm{emo}(\cdot)+\textrm{sim}(\cdot)+\textrm{flu}(\cdot)$ and a
comparison of the \atin model with different objective functions in
supplementary material upon acceptance of this paper.  Here, in
Figure~\ref{tab:outputExamples}, we focus on a discussion of those
cases which we consider particularly difficult, though common in
everyday communication of emotions. In the selection of these
examples, we follow the emotion component model of
\newcite{Scherer2005} and use two examples, which correspond to a
direct (explicit) communication of a subjective feeling (Ex, ID 1, 2),
the description of a bodily reaction (BR, ID 3, 4), and a description
of an event for which an emotion is developed based on a cognitive
appraisal (Ap, ID 5, 6).

The examples which communicate an emotion directly are challenging
because there is no other content available than the emotion that is
described (ID 1, 2). The model has the choice to exchange two out of
three words, and in nearly all cases, it choses to keep ``i'' and
replaces the verb and the emotion word. While the latter is replaced
appropriately, the verb is in most cases not substituted in a
grammatically correct way. We see here that the emotion classification
component in the objective function outrules the language model. This
illustrates one fundamental issue with presumably all existing
affect-related style transfer method: the original emotion is turned into 
the target emotion, but their intensities do not correspond.

In the examples which describe a bodily reaction (ID 3, 4), we see
that the attention mechanism does not allow the words ``over my face''
or ``trembling''
to change. Instead, it finds the other words more likely to be
substituted -- the classifier is not informed about the
meaning of ``trembling'' and ``over my face''. The substituted words
make sense, but content and fluency are
sacrificed again for the maximal emotion intensity available.

Similarly, the emotion classifier and therefore the associated
attention mechanism do not find ``close to the street'' to be
relevant to develop an emotion (ID 5). Instead, other words are
exchanged to introduce the target emotion. These issues
are mostly due to issues in the emotion classification
module. Further, we see that the substitution and selection elements
might have a higher chance to perform well if they considered phrases
instead of isolated words.

We observe a lack of fluency in many of our output sentences, which we
attribute to a dominance of the emotion classifier score. Adapting the
weights of the scores in the objective might have potential, however,
our findings might suggest that content, emotion and fluency are in
conflict with each other -- and that obtaining a particular emotion is
only possible by sacrificing content similarity. Not doing so seems to
lead to non-realistic utterances.

\section{Conclusion \& Future Work}
With this paper, we introduced the task of emotion style transfer,
which we have seen to be particularly difficult, on the one side due
to being on the fence between content and style, and on the other side
due to being a non-binary problem.  Our quantitative analyses have
shown that there is indeed a trade-of between content preservation and
obtaining a target style and that emotion transfer is especially
challenging when the text consists of descriptions of emotions in
which the separation between content and style is not linguistically
clear (as in ``I am happy that X happened''). We propose that such
test sentences based on descriptions of bodily reactions and event
appraisal will be part of future test suits for emotion style
transfer, in order to ensure that this task does not work well only on
particular expressions of emotions.

We identified the challenge to find the right trade-of between
fluency, target emotion, and content preservation. This is
particularly challenging, as it would be desirable to separate the
emotion intensity from our objective function. We therefore propose
that intensity is handled as a fourth component in future work. This
could be combined with a decoder as suggested by
\cite{li_delete_2018}.  Finally, a larger-scale human evaluation
should be carried out to clarify the contribution of each component.

\end{document}